\documentclass{article} 
\usepackage{nips15submit, times}
\usepackage{hyperref}
\usepackage{url}
\usepackage{amsmath}
\usepackage{algorithm}
\usepackage[noend]{algpseudocode}
\usepackage{graphicx}
\usepackage{amssymb}
\usepackage{eqnarray}
\usepackage{multicol}
\usepackage[margin={.75in,1.25in}]{geometry}

\usepackage[backend=biber]{biblatex}
\title{Exploring Alternatives to Softmax Function}

\author{
\large{Kunal Banerjee, Vishak Prasad C, Rishi Raj Gupta$^*$, Karthik Vyas$^*$, Anushree H$^*$, Biswajit Mishra}\\
\texttt{\{kunal.banerjee, vishak.prasad.c, biswajit.mishra\}@intel.com}\\
\large{Intel Corporation}\\
\large{Bangalore, Karnataka, India}
}

\nipsfinalcopy 

\begin{document}
\maketitle
\let\thefootnote\relax\footnote{$^*$Work done during internship at Intel Corporation.}

\begin{abstract}
Softmax function is widely used in artificial neural networks for multiclass classification,
multilabel classification, attention mechanisms, etc.
However, its efficacy is often questioned in literature.
The log-softmax loss has been shown to belong to a more generic class of loss functions, called \textit{spherical family}, and its member log-Taylor softmax loss is arguably the best alternative in this class.
In another approach which tries to enhance the discriminative nature of the softmax function, \textit{soft-margin softmax (SM-softmax)} has been proposed to be the most suitable alternative.
In this work, we investigate Taylor softmax, SM-softmax and our proposed SM-Taylor softmax, an amalgamation of the earlier two functions, as alternatives to softmax function.
Furthermore, we explore the effect of expanding Taylor softmax up to ten terms (original work proposed expanding only to two terms) along with the ramifications of considering Taylor softmax to be a finite or infinite series during backpropagation.
Our experiments for the image classification task on different datasets reveal that there is always a configuration of the SM-Taylor softmax function that outperforms the normal softmax function and its other alternatives.
\end{abstract}

\textbf{Keywords:} softmax, spherical loss, function approximation, classification.

\section{Introduction}\label{sec:intro}
Softmax function is a popular choice in deep learning classification tasks, where it typically appears as the last layer.
Recently, this function has found application in other operations as well, such as the attention mechanisms~\cite{attention}.
However, the softmax function has often been scrutinized in search of finding a better alternative~\cite{sphericalLoss,taylorSoftmax,largeMargin,smSoftmax,dropmax}.

Specifically, Vincent et al. explore the spherical loss family in~\cite{sphericalLoss} that has log-softmax loss as one of its members.
Brebisson and Vincent further work on this family of loss functions and propose log-Taylor softmax as a superior alternative than others, including original log-softmax loss, in~\cite{taylorSoftmax}.

Liu et al. take a different approach to enhance the softmax function by exploring alternatives which may improve the discriminative property of the final layer as reported in~\cite{largeMargin}.
The authors propose \textit{large-margin softmax (LM-softmax)} that tries to increase inter-class separation and decrease intra-class separation.
LM-softmax is shown to outperform softmax in image classification task across various datasets.
This approach is further investigated by Liang et al. in~\cite{smSoftmax}, where they propose \textit{soft-margin softmax (SM-softmax)} that provides a finer control over the inter-class separation compared to LM-softmax.
Consequently, SM-softmax is shown to be a better alternative than its predecessor LM-softmax~\cite{smSoftmax}.

In this work, we explore the various alternatives proposed for softmax function in the existing literature. 
Specifically, we focus on two contrasting approaches based on spherical loss and discriminative property and choose the best alternative that each has to offer: log-Taylor softmax loss and SM-softmax, respectively.  
Moreover, we enhance these functions to investigate whether further improvements can be achieved.
The contributions of this paper are as follows:
\begin{itemize}
\item We propose SM-Taylor softmax -- an amalgamation of Taylor softmax and SM-softmax.
\item We explore the effect of expanding Taylor softmax up to ten terms (original work~\cite{taylorSoftmax} proposed expanding only to two terms) and we prove higher order even terms in Taylor's series are positive definite, as needed in Taylor softmax.
\item We explore ramifications of considering Taylor softmax to be a finite or infinite series during backpropagation.
\item We compare the above mentioned variants with Taylor softmax, SM-softmax and softmax for image classification task.
\end{itemize}
It may be pertinent to note that we do not explore other alternatives such as, dropmax~\cite{dropmax}, because it requires the true labels to be available; however, such labels may not exist in other tasks where softmax function is used, for example, attention mechanism~\cite{attention}.
Consequently, dropmax cannot be considered as a drop-in replacement for softmax universally and hence we discard it.

The paper is organized as follows.
Section~\ref{sec:alternate} elaborates on the softmax function and its several alternatives explored here.
Experimental results are provided in Section~\ref{sec:result}.
Section~\ref{sec:concl} concludes the paper and shares our plan for future work.

\section{Alternatives to softmax}\label{sec:alternate}
In this section, we provide a brief overview of the softmax function and its alternatives explored in this work.

\subsection{Softmax}
The softmax function $sm:\mathbb{R}^{K}\to \mathbb{R}^{K}$ is defined by the formula:
\begin{equation}\label{eq:1}
sm(\mathbf{z})_{i}={\frac{e^{z_{i}}}{\sum_{j=1}^{K}e^{z_{j}}}}{\text{ for }}i=1,\ldots,K{\text{ and }}\mathbf {z} =(z_{1},\dotsc ,z_{K})\in \mathbb{R}^{K}
\end{equation}
To clarify, the exponential function is applied to each element $z_{i}$ of the input vector $\mathbf{z}$ and the resulting values are normalized by dividing by the sum of all the exponentials.
The normalization guarantees that the elements of the output vector $sm(\mathbf{z})$ sum up to 1.

\subsection{Taylor softmax}
The Taylor softmax function as proposed by Vincent et al.~\cite{sphericalLoss} uses second order Taylor series approximation for $e^z$ as $1+z+0.5z^2$. They then derive the Taylor softmax as follows:
\begin{equation}\label{eq:2}
Tsm(\mathbf{z})_{i} = {}  {\frac{1+z_i+0.5z_i^2} {\sum_{j=1}^{K}{1+z_j+0.5z_j^2}} }
 \text{ for } i=1,\ldots,K{\text{ and }}\mathbf {z} =(z_{1},\dotsc ,z_{K})\in \mathbb{R}^{K}
\end{equation}
Moreover, the second order approximation of $e^z$ as $1+z+0.5z^2$ is positive definite, and hence it is suitable to represent a probability distribution of classes~\cite{taylorSoftmax}.
Again, it has a minimum value of 0.5, so the numerator of the equation \ref{eq:2} never becomes zero, that enhances numerical stability.

We explore higher order Taylor series approximation of $e^z$ (as $f^n(z)$) to come up with an $n^{th}$ order Taylor softmax.
\begin{equation}\label{eq:3}
{f^{n}(z)} = {} \sum_{i=0}^{n} \frac{z^{i}}{i!}
\end{equation}
Thus, the Taylor softmax for order $n$ is
\begin{equation}\label{eq:4}
 Tsm^{n}(\mathbf{z})_{i} = {}{\frac{f^{n}(z_{i})} {\sum_{j=1}^{K}{f^{n}(z_{j})}} }
\text{ for } i=1,\ldots,K{\text{ and }}\mathbf {z} =(z_{1},\dotsc,z_{K})\in \mathbb{R}^{K}
\end{equation}
It is important to note that $f^n(z)$ is always positive definite if $n$ is even.
We will prove it by the method of induction.\\
\textit{Base case:} We have already shown that $f^n(z)$ is positive definite for $n=2$ in Section~\ref{sec:alternate}.\\
\textit{Induction hypothesis:} $f^n(z)$ is positive definite for $n=2k$.\\
\textit{Induction step:} We will prove that it holds for $n = 2(k+1) = 2k + 2$, where $k$ is an integer starting from 1.\\ 
We denote $f^{2k+2}(z) = S(k+1)$, so 
\begin{equation*}\label{eq:4}
S(k+1) =
\sum_{i=0}^{2k+2} \frac{z^{i}}{i!}
\end{equation*}
\begin{equation*}\label{eq:5}
\begin{aligned}
S(k+1) = \sum_{i=0}^{2k} \frac{z^{i}}{i!} +
\frac{z^{2k+1}}{(2k+1)!}+ \frac{z^{2k+2}}{(2k+2)!}\\
\text{Let us consider this series with } p \in \mathbb{R} \text{ and }  p > 1\\
S(k+1,p)
= \sum_{i=0}^{2k} \frac{z^{i}}{i!} +
\frac{z^{2k+1}}{(2k+1)!}+ \frac{z^{2k+2}}{(2k+2)!p}\\
\text{Clearly, } S(k+1) > S(k+1,p) \text{  and  }\\
S(k+1,p)= \sum_{i=0}^{2k-1} \frac{z^{i}}{i!} +
\frac{z^{2k}}{(2k)!}(
\frac{(4-p)k+2-p}{2(2k+1)} +\frac{(z+k+1)^{2}}{(2k+1)(2k+2)})\\
> 
\sum_{i=0}^{2k-1} \frac{z^{i}}{i!} +
\frac{z^{2k}}{(2k)!}(
\frac{(4-p)k+2-p}{2(2k+1)})\\
\text { if we select } p < \frac{4k+2}{k+1} \text { then } 
\frac{(4-p)k+2-p}{2(2k+1)} > 0\\
\text{if we set } q =  \frac{2(2k+1)}{(4-p)k+2-p}  
\text{ then the expression becomes}\\
S(k+1,p) >
\sum_{i=0}^{2k-1} \frac{z^{i}}{i!} +
\frac{z^{2k}}{(2k)!q} = S(k,q)\\
\text{We go further  to prove  S(k,q) < S(k-1,r)}, \text{ it requires}\\
q < \frac{4k-2}{k} 
\Leftrightarrow \frac{2(2k+1)}{(4-p)k+2-p}  < \frac{4k-2}{k}\\
\text{which is true if }  p < \frac{4k+2}{k+1}\\
\text{Hence, }S(k+1) > S(k,p) > S(k -1,q) ... > S(1,t)\\
\text{and } S(1,t) > 0 \text{  for  } t= \frac{5}{3}\\
\text{so, } S(k+1) > 0
\end{aligned}
\end{equation*}
$\blacksquare$

The actual back propagation equation for Taylor softmax cross entropy loss function ($L$) is  
\begin{equation}\label{eq:7}
\frac{\partial{L}}{\partial{z}}_{i} = {}  {\frac{f^{n-1}(z_i)}{\sum_{j=1}^{k}f^{n}(z_j)}}
-  y_{i}\frac{f^{n-1}(z_i)}{f^n(z_i)}
\end{equation}
Instead of using equation~\ref{eq:7}, we used softmax like equation~\ref{eq:8} for backpropagation.
For very large $n$ (i.e., as $n$ tends to infinity), equations~\ref{eq:7} and~\ref{eq:8} are equivalent, we denote this variation as Taylor\_{inf}. This equation~\ref{eq:7} is corresponding to negative log likelihood loss function of the Taylor softmax probabilities with a regularizer R(z) defined by equation~\ref{eq:9}; it is because of the regularization effect this method performs better.
\begin{equation}\label{eq:8}
{\frac{\partial{L}}{\partial{z}}}_{i} = {}  Tsm^{n}(\mathbf{z})_i - y_{i}
\end{equation}
\begin{equation}\label{eq:9}
R(\mathbf{z}) = { \log{ \frac{Tsm(\mathbf{z})}{sm(\mathbf{z})} }}
\end{equation}

\subsection{Soft-margin softmax}
Soft-margin (SM) softmax~\cite{smSoftmax} reduces intra-class distances but enhances inter-class discrimination, by introducing a distance margin into the logits. 
The probability distribution for this, as described in~\cite{smSoftmax}, is as follows:
\begin{equation}\label{eq:6}
SMsm(\mathbf{z})_{i} =\frac{e^{z_{i} -m}}  { {\sum_{j \neq i}^{K}e^{z_{j}}}  + e^{z_{i} -m}} \text{ for } i=1,\ldots,K{\text{ and }}\mathbf {z} =(z_{1},\dotsc ,z_{K})\in \mathbb{R}^{K}
\end{equation}

\subsection{SM-Taylor softmax}
SM-Taylor softmax uses the same formula as given in equation~\ref{eq:6}
while using equation~\ref{eq:3}, for some given order $n$, to approximate $e^z$.

\section{Experimental Results}\label{sec:result}
In this section, we share our results for image classification task on MNIST, CIFAR10 and CIFAR100 datasets, where we experiment on the softmax function and its various alternatives.
Note that our goal was not to reach the state of the art accuracy for each dataset but to compare the influence of each alternative.
Therefore, we restricted ourselves to reasonably sized standard neural network architectures with no ensembling and no data augmentation.

\begin{table}[tbh]
 \centering
 \caption{Topologies for different datasets}
 \label{tab:topology}
\begin{tabular}{| c | c | c |}
\hline
MNIST              & CIFAR10                & CIFAR100            \\
\hline
\{Conv[3x3,64]\}x4 & \{Conv[3x3,32],BN\}x2  & Conv[3x3,384]       \\
MaxPool[2x2,2]     & MaxPool[2x2,1]         & MaxPool[2x2,1],DO   \\
\{Conv[3x3,64]\}x3 & Dropout[0.2]           & Conv[1x1,384]       \\
MaxPool[2x2,2]     & \{Conv[3x3,64],BN\}x2  & Conv[2x2,384]       \\
\{Conv[3x3,64]\}x3 & MaxPool[2x2,1]         & \{Conv[2x2,640]\}x2 \\
MaxPool[2x2,2]     & Dropout[0.3]           & MaxPool[2x2,1],DO   \\
FC[256]            & \{Conv[3x3,128],BN\}x2 & Conv[3x3,640]       \\
FC[10]             & MaxPool[2x2,1]         & \{Conv[2x2,768]\}x3 \\
                   & Dropout[0.4]           & Conv[1x1,768]       \\
                   & FC[128],BN             & \{Conv[2x2,896]\}x2 \\
                   & Dropout[0.5]           & MaxPool[2x2,1],DO   \\
                   & FC[10]                 & Conv[3x3,896]       \\
                   &                        & \{Conv[2x2,1024]\}x2\\
                   &                        & MaxPool[2x2,1],DO   \\
                   &                        & Conv[1x1,1024]      \\
                   &                        & Conv[2x2,1152]      \\
                   &                        & MaxPool[2x2,1],DO   \\
                   &                        & Conv[1x1,1152]      \\
                   &                        & MaxPool[2x2,1],DO   \\
                   &                        & FC[100]             \\
\hline
\end{tabular}
\end{table}

The topology that we have used for each dataset is given in Table~\ref{tab:topology}.
The topology for MNIST is taken from~\cite{smSoftmax}; we experimented with the topologies for CIFAR10 and CIFAR100 given in~\cite{smSoftmax} as well to make comparison with the earlier work easier -- however, we could not reproduce the accuracies mentioned in~\cite{smSoftmax} with the prescribed neural networks.
Consequently, we adopted the topology for CIFAR10 mentioned in~\cite{cifar10topo} and for CIFAR100, we borrow the topology given in~\cite{cifar100topo}; in both cases, no data augmentation was applied.
The abbreviations used in Table~\ref{tab:topology} are explained below:
(i) Conv[MxN,K] -- convolution layer with kernel size MxN and K output channels, we always use stride of 1 and padding as ``same'' for convolutions;
(ii) MaxPool[MxN,S] -- maxpool layer with kernel size MxN and stride of S;
(iii) FC[K] -- fullyconnected layer with K output channels, an appropriate \textit{flatten} operation is invoked before the fullyconnected layer that we have omitted for brevity;
(iv) BN -- batchnorm layer with default initialization values;
(v) Dropout[R] -- dropout layer with dropout rate R;
(vi) DO -- dropout layer with rate 0.5, note that CIFAR100 topology uses a uniform rate for all its dropouts;
(vii) \{layer1[,layer2]\}xN -- this combination of layer(s) is repeated N times.
In all these topologies, we replace the final softmax function by each of its alternatives in our experiments.

\begin{table*}[tbh]
 \centering
 \caption{SM-softmax accuracies for different datasets}
 \label{tab:sm_softmax}
\begin{tabular}{| l | r | r | r | r | r | r | r | r | r | r |}
\hline
Dataset &     0 &   0.1 &   0.2 &   0.3 &         0.4 &   0.5 &         0.6 &   0.7 &         0.8 &   0.9\\
\hline
MNIST   & 99.46 & 99.42 & 99.45 & 99.48 &       99.52 & 99.46 & {\bf 99.54} & 99.46 & {\bf 99.54} & 99.47\\
CIFAR10 & 87.09 & 87.10 & 87.29 & 87.15 & {\bf 87.33} & 87.30 & {\bf 87.33} & 87.22 &       87.12 & 87.25\\
CIFAR100& 48.28 & 48.03 & 48.06 & 48.11 &       47.82 & 47.68 & {\bf 48.95} & 48.03 &       47.96 & 48.02\\
\hline
\end{tabular}
\end{table*}

Table~\ref{tab:sm_softmax} shows the effect of varying soft margin $m$ on accuracy for the three datasets.
We vary $m$ from 0 to 0.9 with a step size of 0.1, as prescribed in the original work~\cite{smSoftmax}.
We note that $m$ set to 0.6 provided the best accuracy for all the datasets considered, although there are other values which provide the same best accuracy for MNIST and CIFAR10.
Hence, for simplicity, we fix $m$ to 0.6 for all further experiments. 

\begin{table}
  \centering
  \caption{Comparison among softmax and its alternatives}
  \label{tab:result}
\begin{tabular}{| l | l | r | r | r | r | r | r |}
\hline
Dataset	& Variants    & Accuracy    &           2 &     4 &     6 &     8 &    10\\
\hline
MNIST   & softmax     & 99.41       &             &       &       &       &      \\
        & Taylor      & 99.65       & 99.54       & 99.59 & 99.50 & 99.65 & 99.51\\  
	    & Taylor\_inf & 99.62	    & 99.54       & 99.60 & 99.59 & 99.62 & 99.47\\
        & SM-softmax  & 99.54       &             &       &       &       &      \\	
        & SM-Taylor   & {\bf 99.67} & {\bf 99.67} & 99.59 & 99.63 & 99.47 & 99.45\\
\hline
CIFAR10	& softmax     & 86.87       &             &       &       &       &      \\			             
        & Taylor      & 87.29       & 86.86       & 87.06 & 87.17 & 87.29 & 87.29\\
        & Taylor\_inf &	87.46       & 87.46       & 87.37 & 87.34 & 87.00 & 87.38\\
        & SM-softmax  & 87.33       &             &       &       &       &      \\
        & SM-Taylor   & {\bf 87.47} & {\bf 87.47} & 86.86 & 87.08 & 87.08 & 87.27\\
\hline
CIFAR100& softmax     & 48.57       &             &       &       &       &      \\
        & Taylor      & 49.94       & 44.70       & 49.24 & 49.94 & 49.84 & 49.04\\
        & Taylor\_inf &	49.81       & 44.62       & 47.31 & 49.81 & 46.69 & 45.97\\
        & SM-softmax  & 48.95       &             &       &       &       &      \\
        & SM-Taylor   & {\bf 49.95} & 44.77 & {\bf 49.95} & 49.56 & 49.69 & 48.11\\
\hline
\end{tabular}
\end{table}

Table~\ref{tab:result} compares the softmax function and its various alternatives with respect to image classification task for the different datasets.
For Taylor softmax (Taylor) and its variant where we consider equation~\ref{eq:8} from Section~\ref{sec:alternate} while doing gradient calculation (Taylor\_inf), we look into Taylor series expansion of orders 2 to 10 with a step size of 2.
For SM-Taylor softmax, we follow the same expansion orders while keeping soft margin fixed at 0.6.
For these three variants, we choose the order which gives the best accuracy and mention it again in the column labeled ``Accuracy''.
As can be seen from Table~\ref{tab:result}, there is always a configuration for SM-Taylor softmax (namely, $m=0.6,~order=2$ for MNIST and CIFAR10, and $m=0.6,~order=4$ for CIFAR100) that outperforms other alternatives.

\begin{figure}[tbh]
  \centering
  \includegraphics[width=0.5\textwidth]{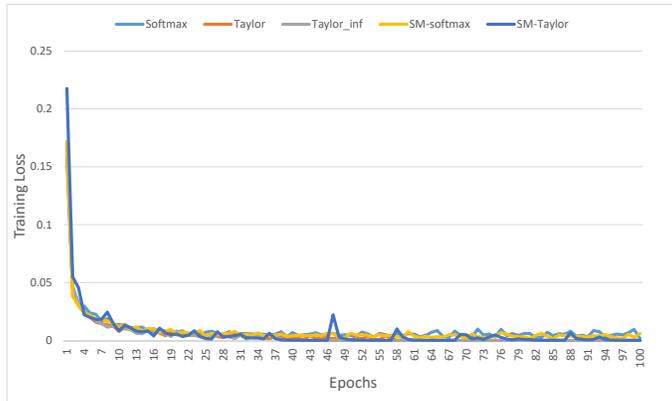}
  \caption{Plot of training loss vs epochs for MNIST dataset.}
  \label{fig:loss_mnist}
\end{figure}
\begin{figure}[tbh]
  \centering
  \includegraphics[width=0.5\textwidth]{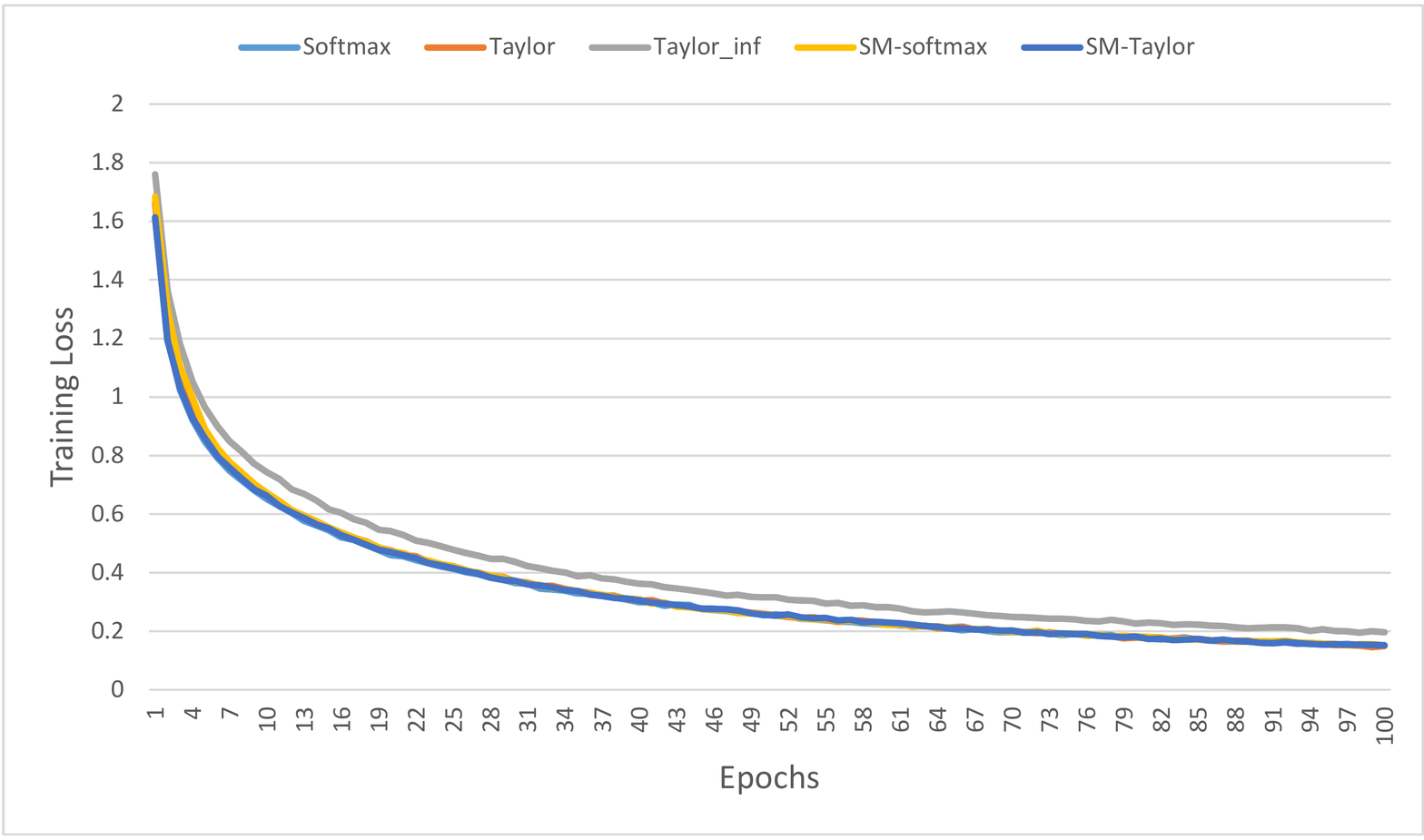}
  \caption{Plot of training loss vs epochs for CIFAR10 dataset.}
  \label{fig:loss_cifar10}
\end{figure}
\begin{figure}[tbh]
  \centering
  \includegraphics[width=0.5\textwidth]{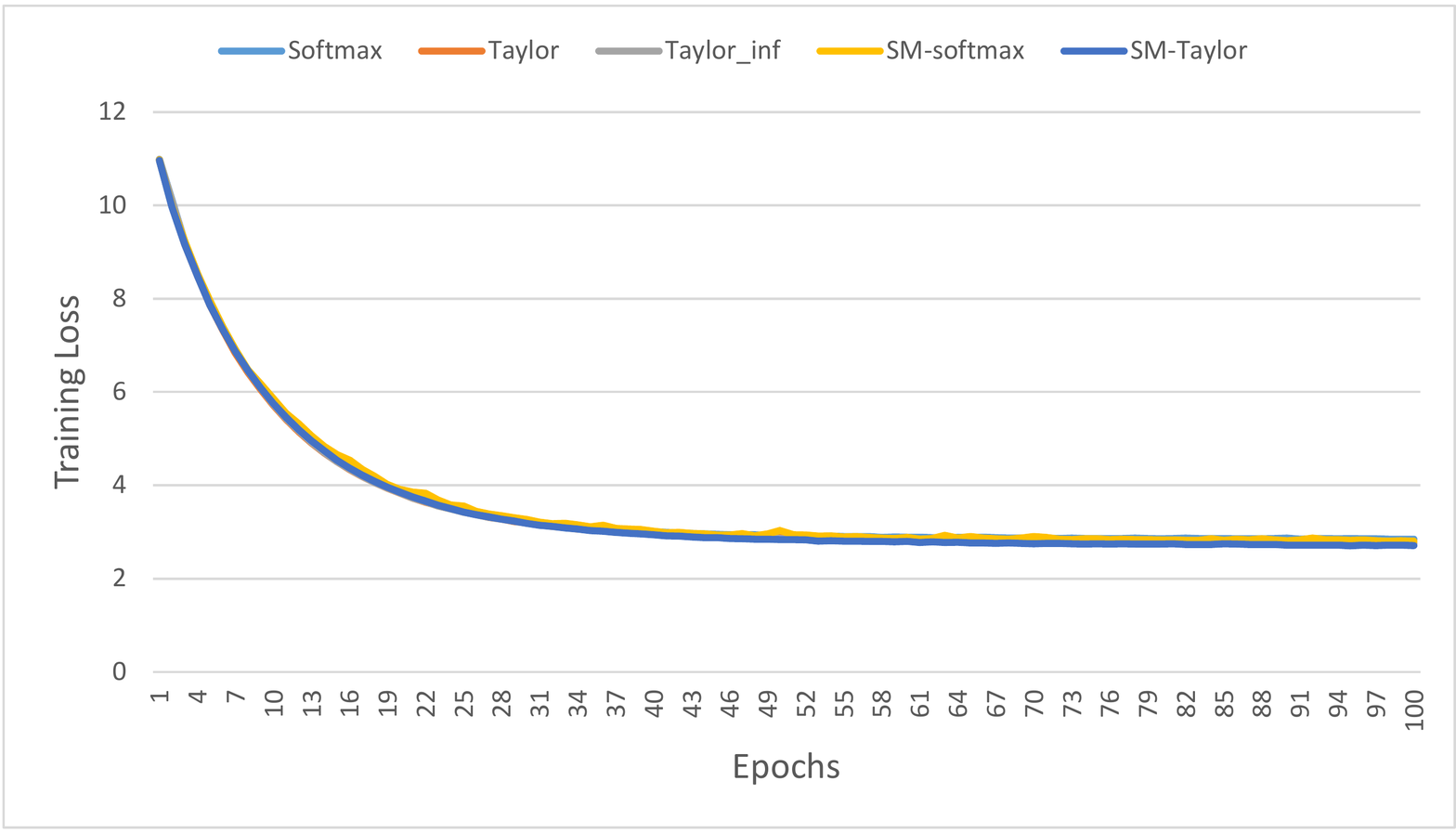}
  \caption{Plot of training loss vs epochs for CIFAR100 dataset.}
  \label{fig:loss_cifar100}
\end{figure}

The plots for training loss vs epochs for MNIST, CIFAR10 and CIFAR100 are given in Figure~\ref{fig:loss_mnist}, Figure~\ref{fig:loss_cifar10} and Figure~\ref{fig:loss_cifar100}, respectively,
It may be pertinent to note that in Figure~\ref{fig:loss_mnist}, we see fluctuation in the training loss for the softmax function, whereas the plot is comparatively smoother for all its alternatives.

\section{Conclusion and Future Work}\label{sec:concl}
Softmax function can be found in almost all modern artificial neural network models whose applications range from image classification, object detection, language translation to many more,
However, there has been a lot of research dedicated to finding a better alternative to this popular softmax function.
One approach explores the loss functions belonging to the spherical family, and proposes log-Taylor softmax loss as arguably the best loss function in this family~\cite{sphericalLoss}
Another approach that tries to amplify the discriminative nature of the softmax function, proposes soft-margin (SM) softmax as the most appropriate alternative.
In this work, we investigate Taylor softmax, soft-margin softmax and our proposed SM-Taylor softmax as alternatives to softmax function.
Moreover, we study the effect of expanding Taylor softmax up to ten terms, in contrast to the original work that expanded only to two terms, along with the ramifications of considering Taylor softmax to be a finite or infinite series during gradient computation.
Through our experiments for the image classification task on different datasets, we establish that there is always a configuration of the SM-Taylor softmax function that outperforms the original softmax function and its other alternatives.

In future, we want to explore bigger models and datasets, especially, the ILSVRC2012 dataset~\cite{ILSVRC15} and its various winning models over the years.
Next we want to explore other tasks where softmax is used, for example, image caption generation~\cite{imageCaption} and language translation~\cite{attention}, and check how well do the softmax alternatives covered in this work perform for the varied tasks.
Ideally, we would like to discover an alternative to softmax that can be considered as its drop-in replacement irrespective of the task at hand.

\printbibliography

@inproceedings{attention,
  author    = {Ashish Vaswani and
               Noam Shazeer and
               Niki Parmar and
               Jakob Uszkoreit and
               Llion Jones and
               Aidan N. Gomez and
               Lukasz Kaiser and
               Illia Polosukhin},
  title     = {Attention is All you Need},
  booktitle = {NeurIPS},
  pages     = {5998--6008},
  year      = {2017},
}

@inproceedings{sphericalLoss,
  author    = {Pascal Vincent and
               Alexandre de Br{\'{e}}bisson and
               Xavier Bouthillier},
  title     = {Efficient Exact Gradient Update for training Deep Networks with Very
               Large Sparse Targets},
  booktitle = {NeurIPS},
  pages     = {1108--1116},
  year      = {2015},
}

@inproceedings{taylorSoftmax,
  author = {Alexandre de Br{\'{e}}bisson and
            Pascal Vincent},
  title = {An Exploration of Softmax Alternatives Belonging to the Spherical Loss Family},
  booktitle = {ICLR},
  year = {2016},
}

@inproceedings{largeMargin,
  author    = {Weiyang Liu and
               Yandong Wen and
               Zhiding Yu and
               Meng Yang},
  title     = {Large-Margin Softmax Loss for Convolutional Neural Networks},
  booktitle = {{ICML}},
  pages     = {507--516},
  year      = {2016},
}

@inproceedings{smSoftmax,
  author    = {Xuezhi Liang and
               Xiaobo Wang and
               Zhen Lei and
               Shengcai Liao and
               Stan Z. Li},
  title     = {Soft-Margin Softmax for Deep Classification},
  booktitle = {{ICONIP}},
  pages     = {413--421},
  year      = {2017},
}

@inproceedings{dropmax,
    author    = {Hae Beom Lee and Juho Lee and Saehoon Kim and Eunho Yang and Sung Ju Hwang},
    title     = {DropMax: Adaptive Variationial Softmax},
    booktitle = {NeurIPS},
    year      = {2018}
}

@misc{cifar10topo,
  author = {Jason Brownlee},
  title = {How to Develop a {CNN} From Scratch for {CIFAR-10} Photo Classification},
  howpublished = {\url{https://machinelearningmastery.com/how-to-develop-a-cnn-from-scratch-for-cifar-10-photo-classification/}},
  note = {Accessed: 2020-06-21}
}

@inproceedings{cifar100topo,
  author    = {Djork{-}Arn{\'{e}} Clevert and
               Thomas Unterthiner and
               Sepp Hochreiter},
  title     = {Fast and Accurate Deep Network Learning by Exponential Linear Units
               (ELUs)},
  booktitle = {{ICLR}},
  year      = {2016},
  url       = {http://arxiv.org/abs/1511.07289},
}

@article{ILSVRC15,
Author = {Olga Russakovsky and Jia Deng and Hao Su and Jonathan Krause and Sanjeev Satheesh and Sean Ma and Zhiheng Huang and Andrej Karpathy and Aditya Khosla and Michael Bernstein and Alexander C. Berg and Li Fei-Fei},
Title = {{ImageNet Large Scale Visual Recognition Challenge}},
Year = {2015},
journal   = {International Journal of Computer Vision (IJCV)},
doi = {10.1007/s11263-015-0816-y},
volume={115},
number={3},
pages={211-252}
}

@inproceedings{imageCaption,
  author    = {Kelvin Xu and
               Jimmy Ba and
               Ryan Kiros and
               Kyunghyun Cho and
               Aaron C. Courville and
               Ruslan Salakhutdinov and
               Richard S. Zemel and
               Yoshua Bengio},
  title     = {Show, Attend and Tell: Neural Image Caption Generation with Visual
               Attention},
  booktitle = {{ICML}},
  series    = {{JMLR} Workshop and Conference Proceedings},
  volume    = {37},
  pages     = {2048--2057},
  year      = {2015},
}
\nocite{*}

\end{document}